# Exploring Emerging Trends and Research Opportunities in Visual Place Recognition

Antonios Gasteratos[1], Konstantinos A. Tsintotas[1,*], Tobias Fischer[2],
Yiannis Aloimonos[3], and Michael Milford[2]

## I. INTRODUCTION

Visual-based recognition, *e.g.*, image classification, object detection, *etc.*, is a long-standing challenge in computer vision and robotics communities. Concerning the roboticists, since the knowledge of the environment is a prerequisite for complex navigation tasks [1], visual place recognition is vital for most localization implementations or re-localization and loop closure detection pipelines within simultaneous localization and mapping (SLAM) [2]. More specifically, it corresponds to the system's ability to identify and match a previously visited location using computer vision tools [3]. Towards developing novel techniques with enhanced accuracy and robustness, while motivated by the success presented in natural language processing methods [4], researchers have recently turned their attention to vision-language models [5], which integrate visual and textual data [6].

With large-scale image-text pairs being almost infinitely available online, such a model is pre-trained to learn the correlations through specific vision-language objectives [7]. This way, rich environmental knowledge is adopted, permitting zero-shot predictions on visual recognition tasks since the outcome relies on the correspondences between the embeddings of the given images and texts [8]. Technically, a text and an image encoder are employed first to extract the corresponding features, while subsequently, the vision-language correlation is learned (see Fig. 1).

This work aims to explore the advantages of this emerging trend when applied to robotic navigation tasks and, in particular, to provide possible research directions for addressing the problem of place recognition. To this end, Section II discusses the potential pathways to pursue in future studies, while the methodological considerations and the expected impact are given in the following sections.

## II. RESEARCH DIRECTIONS

This section proposes three research paths, which mainly correspond to techniques for autonomous platforms that operate in real-world conditions.

[1]Authors are with the Department of Production and Management Engineering, Democritus University of Thrace, 12 Vas. Sophias, GR-671 32, Xanthi, Greece {agaster, ktsintot}@pme.duth.gr
[2]Authors are with the School of Electrical Engineering and Robotics, Queensland University of Technology, 2 George St, QLD 4000, Brisbane, Australia {tobias.fischer, michael.milford}@qut.edu.au
[3]Author is with the Department of Computer Science, University of Maryland, College Park, MD 20742, USA {yiannis}@umiacs.umd.edu
*Corresponding Author

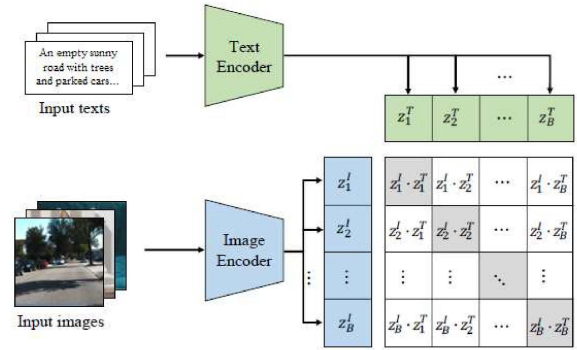

Fig. 1. Image-text contrastive learning. First, a text and an image encoder extract the corresponding features. Next, the vision-language model learns their correlation. For visual-based recognition tasks, the outcome relies on the correspondence between both embeddings extracted from the query.

### A. Semantics via Multimodal Feature Fusion

Integrating visual features from environmental images with textual descriptions can significantly improve robotic navigation systems [9]. In particular, training a vision-language model to learn representations from both modalities jointly permits the creation of a unified feature space that captures the semantic information. For example, by describing a kitchen as "busy", the model can learn detailed semantic representations that help the system differentiate between visually similar but contextually different locations. Incorporating text information provides a richer understanding and leads to robust and accurate localization, especially in environments where visual features alone are insufficient, *e.g.*, sparse surroundings such as long corridors or open fields.

### B. Robustness to Visual Variability

Training a vision-language model on datasets that include images under various lighting conditions, viewpoints, and seasonal changes [10], while simultaneously combining these visual inputs with corresponding textual descriptions can enhance robustness to environments with changing conditions [11], [12]. For example, by learning to integrate and interpret multimodal data, a model trained to understand that a "snow-covered park" and its corresponding version in "summer greenery" are in the exact location ensures more reliable navigation and mapping in dynamic settings where appearance can change significantly over time.

## C. Incremental Learning

As the robot encounters new surroundings or changes in existing ones, applying incremental learning techniques to a vision-language model, *i.e.*, updating it with new data, allows the agent to adapt to new visual and textual information collected during navigation [13]. The model incrementally learns and integrates these updates based on online learning methods, maintaining its relevance and accuracy over time. This adaptability is crucial for long-term reliability in applications with dynamic and evolving environments, such as urban settings or natural landscapes, where conditions can change rapidly.

## III. METHODOLOGICAL CONSIDERATIONS

Formulating vision-language models for place recognition requires a multifaceted strategy that includes data collection, model architecture design, training strategies, and rigorous evaluation protocols.

### A. Data Collection

Firstly, a diverse dataset containing paired visual and textual data from various domains must be curated. Moreover, this should comprise detailed annotations and descriptions to provide rich contextual knowledge.

### B. Model Architecture Design

The model architecture should leverage state-of-the-art techniques, such as transformer-based encoders for visual and textual modalities, while retaining practical tools for multimodal fusion, such as cross-attention layers, to allow the model to learn intricate relationships between visual features and textual descriptions.

### C. Training Strategies

Transfer learning is necessary during the training phase, as pre-trained models on large-scale datasets, *e.g.*, ImageNet for visual data [14], and large text corpora for natural language understanding can be fine-tuned on specific visual place recognition datasets. At the same time, techniques such as contrastive learning can align visual and textual embeddings in a shared latent space, improving the model's ability, especially in contextually similar but visually different scenes.

### D. Evaluation protocol

Evaluation should be comprehensive, involving metrics like precision and recall specifically for loop closure detection tasks, alongside robustness tests against environmental variability [15]. Benchmarking against standard visual place recognition datasets will help gauge the model's practical effectiveness. Computational efficiency is also critical; thus, optimizing the model for real-time inference through techniques like model quantization, pruning, and efficient inference frameworks is necessary to ensure it meets the processing demands of a robotic system.

## IV. CONCLUSION AND EXPECTED IMPACT

Vision-language models are poised to significantly improve place recognizers' accuracy, autonomy, and robustness by harnessing their ability to use the combined power of a scene's visual and textual information. This advancement aims to extend the applicability of robotic platforms to scenarios where traditional methods fail, such as sparse or visually challenging environments. By combining multimodal data or understanding the image's deeper semantic information, such models promise to overcome current limitations. At the same time, new avenues will open for applications in which autonomous systems will be more adaptive, interacting seamlessly with their surroundings for accomplishing complex tasks, *e.g.*, transportation and logistics, surveillance, and exploration.